\documentclass[runningheads]{llncs}
\usepackage[T1]{fontenc}
\usepackage{multirow}
\usepackage[russian,british]{babel}
\usepackage{graphicx}
\begin{document}
\title{Exploring Fine-tuned Generative Models for Keyphrase Selection: A Case Study for Russian}
\titlerunning{Exploring Fine-tuned Generative Models for Keyphrase Selection}
%
%
%

\author{Anna Glazkova\inst{1}\orcidID{0000-0001-8409-6457} \and
Dmitry Morozov\inst{2}\orcidID{0000-0003-4464-1355}}
%
\authorrunning{A. Glazkova, D. Morozov}
%
\institute{
University of Tyumen, Tyumen, Russia\\
\email{a.v.glazkova@utmn.ru}\\\and
Novosibirsk State University, Novosibirsk, Russia\\
\email{morozowdm@gmail.com}}
\maketitle              
\begin{abstract}
Keyphrase selection plays a pivotal role within the domain of scholarly texts, facilitating efficient information retrieval, summarization, and indexing. In this work, we explored how to apply fine-tuned generative transformer-based models to the specific task of keyphrase selection within Russian scientific texts. We experimented with four distinct generative models, such as ruT5, ruGPT, mT5, and mBART, and evaluated their performance in both in-domain and cross-domain settings. The experiments were conducted on the texts of Russian scientific abstracts from four domains: mathematics \& computer science, history, medicine, and linguistics. The use of generative models, namely mBART, led to gains in in-domain performance (up to 4.9\% in BERTScore, 9.0\% in ROUGE-1, and 12.2\% in F1-score) over three keyphrase extraction baselines for the Russian language. Although the results for cross-domain usage were significantly lower, they still demonstrated the capability to surpass baseline performances in several cases, underscoring the promising potential for further exploration and refinement in this research field.

\end{abstract}
\section{Introduction}

Identifying and extracting keyphrases from a document is an essential task in natural language processing aimed at summarizing the crucial information presented in the source document. Keyphrases facilitate retrieval of documents from large text corpora and show their efficacy in various tasks, such as text summarization and content analysis \cite{narin2021content,widyassari2022review}. The two main approaches to keyphrase selection are extracting keyphrases directly from the text and generating keyphrases based on the semantics of the text through its generalization and paraphrasing~\cite{meng2017deep}. In the second case, the task of keyphrase selection is similar to the task of abstractive text summarization \cite{cano2019keyphrase}.

The majority of widely used approaches to keyphrase extraction are based on unsupervised identifying the most significant words and phrases from the text, in particular, \cite{bougouin2013topicrank,campos2020yake}. Although keyphrase extraction algorithms show impressive results across various text corpora, they have a number of limitations. Specifically, they cannot generate keyphrases that are not explicitly stated in the text. In practice, the list of keyphrases for a news article or scientific paper may comprise both keyphrases present in the text and keyphrases related to the content of the text but not explicitly mentioned in it. This limitation can be overcome by deep neural models, particularly by pre-trained language models for text generation.

In this paper, we fine-tune several text generation models for selecting keyphrases for Russian scientific texts. We compare four models, including ruT5, ruGPT, mT5, and mBART, and perform their in-domain and cross-domain evaluation in terms of three different metrics. In this study, we focus on fine-tuned models since it has been widely demonstrated that fine-tuning language models is effective across a broad spectrum of natural language processing tasks. 

We aim to answer the following research questions.
\begin{enumerate}
    \item How do text generation models perform compared to known baselines for selecting keyphrases in Russian texts?
    \item How does the performance of keyphrase generation for Russian differ between in-domain and cross-domain settings?
\end{enumerate}

The rest of the paper is organized as follows. Section \ref{related} contains a brief review of related work. Section \ref{method} describes the datasets and models utilized in this study and provides an experimental setup. Section \ref{results} discusses the results. Section \ref{conclusion} concludes this paper.

\section{Related Work}\label{related}
\subsection{Recent Advances in Keyphrase Selection}

Most widely used approaches to keyphrase selection are based on identifying the most significant words and phrases that give meaning to text content using unsupervised learning principles. In this case, the task is commonly referred to as keyphrase extraction \cite{papagiannopoulou2020review}. In particular, keyphrase extraction approaches cover statistic-based methods, such as RAKE \cite{rose2010automatic} and YAKE! \cite{campos2020yake}, and graph methods, such as TopicRank \cite{bougouin2013topicrank}. Some of the keyphrase extraction methods belong to supervised approaches, for example, KEA \cite{witten1999kea}), which is based on the Naïve Bayes classifier.

Another possible problem statement of keyphrase selection is keyphrase generation. In contrast to keyphrase extraction, generative approaches can produce keyphrases that are absent in a source text \cite{ccano2019keyphrase}. The authors of \cite{meng2017deep} proposed the CopyRNN model that comprises an encoder, which forms a hidden representation of the source text, and a decoder, which produces keyphrases using that representation. Later, more powerful architectures were proposed \cite{chen2018keyphrase,zhang2018keyphrase}. The scholars also experimented with various training paradigms, including reinforcement learning \cite{chan2019neural} and adversarial training \cite{swaminathan2020preliminary}.

The rise of pre-trained language models \cite{devlin-etal-2019-bert,liu2019roberta} has led to notable changes in various natural language processing tasks, particularly in how to use and benefit from these models for specific tasks. Recent studies applied state-of-the-art transformer-based models to both tasks of keyphrase extraction and generation. Some studies proposed unsupervised approaches and focused on embedding-based models \cite{grootendorst2020keybert,sahrawat2020keyphrase}, which utilized pre-trained contextualized embeddings for extracting keyphrases. The authors of the survey \cite{song2023survey} noted that supervised transformer-based approaches to keyphrase extraction typically integrate candidate keyphrase extraction and their importance estimation within an end-to-end learning framework \cite{song2022hyperbolic,sun2020joint}. In \cite{mu2020keyphrase} the task of keyphrase extraction is considered as a sequence labelling task. In the field of keyphrase generation, great success was achieved using transformer language models pre-trained with different pre-training objectives \cite{kulkarni2022learning,bird2006nltk}. In particular, the researchers often use BART \cite{lewis2020bart} and T5 \cite{raffel2020exploring} for generating keyphrases \cite{chen2023enhancing,jiang2023generating,kulkarni2022learning}. In \cite{wu2022pre}, the authors presented the in-domain BART models for scientific and news domains. Several preliminary studies are devoted to instruction-based keyphrase generation using large language models, namely, ChatGPT \cite{martinez2023chatgpt,song2023chatgpt}.

\subsection{Keyphrase Selection for Russian Texts}

A number of studies have investigated the use of unsupervised algorithms for keyphrase extraction from Russian texts. In particular, the authors of \cite{sandul2018keyword,sokolova2018keyphrase} utilized the RAKE algorithm for analyzing Russian texts from various domains. The paper \cite{mitrofanova2022experiments} provides a comparison of several algorithms, including TF-IDF, YAKE!, RAKE, KeyBERT, and others, on a set of heterogeneous documents containing news, scientific, and literary texts. The work \cite{guseva2024} presents a large-scale comparison of keyphrase extraction approaches for Russian popular science texts. In \cite{morozov2023generation}, the dataset for keyphrase selection from Russian texts from mathematical and computer science domains is presented. The authors also compare several common unsupervised approaches for keyphrase extraction. In~\cite{wienecke2020automatic}, the keyphrases for Russian scientific texts are determined using a Latent Dirichlet allocation topic model. 

To date, some studies explored supervised approaches for selecting keyphrases for Russian texts. The papers \cite{sokolova2018keyphrase,sokolova2017automatic} investigate the effectiveness of the KEA algorithm for Russian texts. The paper \cite{koloski2021extending} proposes an approach combining traditional unsupervised algorithms and neural networks. The approach is evaluated on a multilingual corpus, including Russian-language texts. In \cite{nguyen2021keyphrase}, the authors present a neural model for keyphrase extraction that calculates features from traditional statistical metrics and new state-of-the-art sentence embeddings. The authors of \cite{glazkova2023keyphrase} fine-tune a multilingual text-to-text transformer~(mT5) \cite{xue2021mt5} for generating keyphrases for scientific texts from mathematical and computer science domains. For some metrics, the fine-tuned model outperformed unsupervised baselines.
\newline

A brief review of the related work has shown that pre-trained language models have great potential for the task of keyphrase selection. The majority of existing research on keyphrase generation is conducted on English text corpora. However, for other languages, the task of keyphrase generation has been less explored. Despite this, there is a large number of non-English online sources that also require automatic analysis and systematization. This work aims to address the existing research gap in the investigation of state-of-the-art language models for keyphrase generation for Russian texts.

\section{Method}\label{method}

To answer research questions, we collected scientific texts and keyphrases for four different domains. The texts were divided into train and test sets. We fine-tuned several transformer-based models for generating keyphrases using train sets. The results of generative models were compared with the results of common baselines for keyphrase extraction in terms of three evaluation metrics.

\subsection{Data}

We utilized the Math\&CS dataset \cite{morozov2022dataset} that consists of abstracts and their corresponding keyphrases collected from the online resources MathNet and Cyberleninka and described in \cite{morozov2023generation}. Math\&CS contains texts from mathematical and computer science domains. To perform cross-domain evaluation, we collected 22500 pairs of abstracts and corresponding keyphrases from Cyberleninka for three domains, namely, historical, medical, and linguistic. 

The characteristics of data are presented in Table \ref{table_data}. The average numbers of tokens and sentences are defined using the NLTK package \cite{bird2006nltk}. The percentage of absent keyphrases means the proportion of keyphrases from the list of keyphrases that do not appear in the corresponding abstract text.

\begin{table}[]
\centering
\caption{Data statistics}
\begin{tabular}{|p{3.7cm}|l|l|l|l|}
\hline
Characteristic & Math\&CS & Historical & Medical & Linguistic \\ \hline
Train size & 5844 & 5000 & 5000 & 5000 \\ \hline
Test size & 2504 & 2500 & 2500 & 2500 \\ \hline
Avg number of sentences & 3.73$\pm$2.75 & 2.93$\pm$2.03 & 6.58$\pm$4.91 & 4.57$\pm$3.03 \\ \hline
Avg number of tokens & 74.16$\pm$61.65 & 58.04$\pm$35.99 & 141.94$\pm$99.13 & 100.43$\pm$68.57 \\ \hline
Avg number of keyphrases per text & 4.34$\pm$1.5 & 4.8$\pm$1.81 & 4.07$\pm$1.35 & 4.97$\pm$1.62 \\ \hline
Absent keyphrases, \% & 53.66 & 60.47 & 40.13 & 53.16 \\ \hline
\end{tabular}
\label{table_data}
\end{table}

\subsection{Models}

We used four pre-trained transformer-based models. The list of the models and their parameters are given in Table \ref{table_models}. 

The ruGPT model was fine-tuned with a causal language modeling objective with a maximum sequence length of 1024 tokens for three epochs. We used the learning rate of 4e-5 and the Adam optimizer with $\beta$1 = 0.9, $\beta$2 = 0.999, and $\epsilon$ = 1e-8. The input text for ruGPT was presented as follows: \textit{text} + $<|keyphrases|>$ + \textit{list of keyphrases} + $<|end|>$. $<|keyphrases|>$ and $<|end|>$ are special tokens. The list of keyphrases represented a string of keyphrases divided by commas. To generate keyphrases for the test set, each text from the test set was supplemented by a special token $<|keyphrases|>$.

The mT5, ruT5, and mBART models were fine-tuned for ten epochs with a maximum sequence length of 256 tokens. The learning rate was 1e-5 for ruT5 and 4e-5 for mT5 and mBART. The optimizer was the same as for ruGPT. To generate keyphrases for the test set, the model input was only the text of the abstract.

For all generative models, we did not limit the number of generated keyphrases. The optimal number of keyphrases for each text was defined by the models themselves.

\begin{table}[h]
\centering
\caption{Model overview}
\begin{tabular}{|p{3.8cm}|p{2.4cm}|p{1.3cm}|p{4.3cm}|}
\hline
\textbf{Model} & \textbf{Architecture} & \textbf{Params} & \textbf{Data source} \\
\hline
ruT5 (ruT5-base) \cite{zmitrovich2023family} & encoder-decoder & 222M & \multirow{2}{*}{\begin{tabular}[c]{@{}l@{}}Wikipedia, news texts,\\ Librusec, C4, OpenSubtitles\end{tabular}} 
\\ \cline{1-3} 
ruGPT (rugpt-3-medium) \cite{zmitrovich2023family} & decoder-only & 355M &  \\
\hline
mT5 (mT5-base) \cite{xue2021mt5} & encoder-decoder & 580M & Common Crawl (mC4) \\
\hline
mBART (mbart-large-50) \cite{tang2020multilingual} & encoder-decoder & 680M & Common Crawl (CC25) + XLMR \\
\hline
\end{tabular}
\label{table_models}
\end{table}

\subsection{Baselines}

We compared the results of generative models with the results of the following baselines.

\begin{itemize}
    \item RuTermExtract \cite{rutermextract}, a package that determines important terms within a given piece of content using PyMorphy2 \cite{korobov2015morphological} for morphological analysis.
    \item YAKE! \cite{campos2020yake}, an unsupervised method leveraging statistical attributes of text to select the most relevant keyphrases. We used the implementation of YAKE! provided by keyphrases.mca.nsu.ru via API.
    \item KeyBERT \cite{grootendorst2020keybert}, a method employing document and word embeddings generated by BERT \cite{devlin-etal-2019-bert}, along with cosine similarity, to identify the sub-phrases within a document that closely resemble the document as a whole. For KeyBERT, we utilized ruBERT-base-cased \cite{kuratov2019adaptation} as a basic model.
\end{itemize}

\subsection{Evaluation Metrics}

To evaluate the results, we used the following metrics: BERTScore \cite{zhangbertscore}, ROUGE-1 \cite{lin2004rouge}, and the full-match F1-score. We have chosen several diverse metrics since the issue of assessing the quality of text generation remains challenging \cite{celikyilmaz2020evaluation}. Existing approaches to assessing the quality of the selection of keyphrases also evaluate different aspects of the list of keyphrases, such as semantic similarity, similarity of n-grams, or a complete coincidence of generated keyphrases with keywords composed by a human expert \cite{papagiannopoulou2020review}.

BERTScore uses contextual embeddings pre-trained by BERT-based models and matches words in the source and generated texts in terms of cosine similarity. The ROUGE-1 score calculates the number of matching unigrams between the model-generated text and the reference. To calculate ROUGE-1 and BERTScore, the keyphrases for each text were combined into a string with a comma as a separator. The full-match F1-score assesses the number of exact matches between the original and produced sets of keyphrases, computed as the harmonic mean of precision and recall. To calculate F1-score and ROUGE-1, the keyphrases were preliminarily lemmatized to reduce the number of mismatches.

\section{Results and Discussion}\label{results}

To answer \textbf{RQ1}, we compared the results of baselines and generative models. The scores are given in Tables \ref{table_baseline_results} and \ref{table_in_domain_results}. For each baseline model, we calculated BERTScore, ROUGE-1, and F1-score at the top 5, 10, and 15 keyphrases. 

\begin{table}[h!]
\centering
\caption{Baseline results, \%. BERTScore -- BS, ROUGE-1 - R1, F1-score -- F1}
\begin{tabular}{|lllll|}
\hline
\multicolumn{5}{|c|}{\textbf{RuTermExtract}} \\ \hline\hline
\multicolumn{1}{|l|}{Metric} & \multicolumn{1}{l|}{Math\&CS} & \multicolumn{1}{l|}{Historical} & \multicolumn{1}{l|}{Medical} & Linguistic \\ \hline\hline
\multicolumn{1}{|l|}{BS@5} & \multicolumn{1}{l|}{75.85} & \multicolumn{1}{l|}{73.47} & \multicolumn{1}{l|}{\textbf{74.93}} & 74.04 \\ \hline
\multicolumn{1}{|l|}{BS@10} & \multicolumn{1}{l|}{\textbf{75.95}} & \multicolumn{1}{l|}{\textbf{73.53}} & \multicolumn{1}{l|}{74.82} & \textbf{74.15} \\ \hline
\multicolumn{1}{|l|}{BS@15} & \multicolumn{1}{l|}{75.86} & \multicolumn{1}{l|}{73.32} & \multicolumn{1}{l|}{74.5} & 73.97 \\ \hline\hline
\multicolumn{1}{|l|}{R1@5} & \multicolumn{1}{l|}{25.77} & \multicolumn{1}{l|}{19.58} & \multicolumn{1}{l|}{\textbf{22.86}} & 22.51 \\ \hline
\multicolumn{1}{|l|}{R1@10} & \multicolumn{1}{l|}{\textbf{26.49}} & \multicolumn{1}{l|}{\textbf{20.54}} & \multicolumn{1}{l|}{22.65} & \textbf{23.09} \\ \hline
\multicolumn{1}{|l|}{R1@15} & \multicolumn{1}{l|}{25.82} & \multicolumn{1}{l|}{19.95} & \multicolumn{1}{l|}{21.53} & 21.95 \\ \hline\hline
\multicolumn{1}{|l|}{F1@5} & \multicolumn{1}{l|}{9.75} & \multicolumn{1}{l|}{8.48} & \multicolumn{1}{l|}{\textbf{11.35}} & 10.1 \\ \hline
\multicolumn{1}{|l|}{F1@10} & \multicolumn{1}{l|}{\textbf{11.02}} & \multicolumn{1}{l|}{\textbf{9.65}} & \multicolumn{1}{l|}{11.2} & \textbf{11.19} \\ \hline
\multicolumn{1}{|l|}{F1@15} & \multicolumn{1}{l|}{10.86} & \multicolumn{1}{l|}{9.54} & \multicolumn{1}{l|}{10.85} & 10.95 \\ \hline\hline

\multicolumn{5}{|c|}{\textbf{YAKE!}} \\ \hline\hline
\multicolumn{1}{|l|}{BS@5} & \multicolumn{1}{l|}{69.13} & \multicolumn{1}{l|}{66.5} & \multicolumn{1}{l|}{68.76} & 66.5 \\ \hline
\multicolumn{1}{|l|}{BS@10} & \multicolumn{1}{l|}{68.08} & \multicolumn{1}{l|}{65.26} & \multicolumn{1}{l|}{68.28} & 65.78 \\ \hline
\multicolumn{1}{|l|}{BS@15} & \multicolumn{1}{l|}{66.81} & \multicolumn{1}{l|}{64.34} & \multicolumn{1}{l|}{67.26} & 65.44 \\ \hline\hline
\multicolumn{1}{|l|}{R1@5} & \multicolumn{1}{l|}{17.34} & \multicolumn{1}{l|}{13.24} & \multicolumn{1}{l|}{16.2} & 12.78 \\ \hline
\multicolumn{1}{|l|}{R1@10} & \multicolumn{1}{l|}{20.76} & \multicolumn{1}{l|}{16.03} & \multicolumn{1}{l|}{19.97} & 16.9 \\ \hline
\multicolumn{1}{|l|}{R1@15} & \multicolumn{1}{l|}{20.87} & \multicolumn{1}{l|}{16.31} & \multicolumn{1}{l|}{20.21} & 17.71 \\ \hline\hline
\multicolumn{1}{|l|}{F1@5} & \multicolumn{1}{l|}{2.67} & \multicolumn{1}{l|}{2.75} & \multicolumn{1}{l|}{4.48} & 1.67 \\ \hline
\multicolumn{1}{|l|}{F1@10} & \multicolumn{1}{l|}{5.04} & \multicolumn{1}{l|}{4.26} & \multicolumn{1}{l|}{6.29} & 4.84 \\ \hline
\multicolumn{1}{|l|}{F1@15} & \multicolumn{1}{l|}{6.03} & \multicolumn{1}{l|}{5.32} & \multicolumn{1}{l|}{6.91} & 5.88 \\ \hline\hline
\multicolumn{5}{|c|}{\textbf{KeyBERT}} \\ \hline\hline
\multicolumn{1}{|l|}{BS@5} & \multicolumn{1}{l|}{68.87} & \multicolumn{1}{l|}{66.74} & \multicolumn{1}{l|}{67.95} & 68.77 \\ \hline
\multicolumn{1}{|l|}{BS@10} & \multicolumn{1}{l|}{67.95} & \multicolumn{1}{l|}{65.65} & \multicolumn{1}{l|}{67.23} & 67.89 \\ \hline
\multicolumn{1}{|l|}{BS@15} & \multicolumn{1}{l|}{66.98} & \multicolumn{1}{l|}{64.71} & \multicolumn{1}{l|}{66.24} & 66.99 \\ \hline\hline
\multicolumn{1}{|l|}{R1@5} & \multicolumn{1}{l|}{15.38} & \multicolumn{1}{l|}{12.24} & \multicolumn{1}{l|}{10.9} & 15.64 \\ \hline
\multicolumn{1}{|l|}{R1@10} & \multicolumn{1}{l|}{16.97} & \multicolumn{1}{l|}{13.61} & \multicolumn{1}{l|}{12.13} & 16.69 \\ \hline
\multicolumn{1}{|l|}{R1@15} & \multicolumn{1}{l|}{17.68} & \multicolumn{1}{l|}{13.95} & \multicolumn{1}{l|}{12.32} & 16.46 \\ \hline\hline
\multicolumn{1}{|l|}{F1@5} & \multicolumn{1}{l|}{2.2} & \multicolumn{1}{l|}{1.73} & \multicolumn{1}{l|}{1.77} & 1.99 \\ \hline
\multicolumn{1}{|l|}{F1@10} & \multicolumn{1}{l|}{2.76} & \multicolumn{1}{l|}{2.47} & \multicolumn{1}{l|}{2.43} & 2.74 \\ \hline
\multicolumn{1}{|l|}{F1@15} & \multicolumn{1}{l|}{3.31} & \multicolumn{1}{l|}{2.93} & \multicolumn{1}{l|}{2.63} & 3.11 \\ \hline
\end{tabular}
\label{table_baseline_results}
\end{table}

Among baselines, the best results for test data were demonstrated by RuTermExtract. The highest BERTScore and ROUGE-1 were obtained for Math\&CS (75.97\% and 26.49\%). The highest F1-score was achieved for the medical domain (11.35\%). Other scores are provided in Table \ref{table_baseline_results}. The best results for each domain are shown in bold.

Table \ref{table_in_domain_results} demonstrates the in-domain results of generative models. In this case, the models were fine-tuned and tested on the same domain. The values that outperformed the corresponding scores achieved by baselines are shown in bold. The best results for each domain across all models are underlined. In most cases, generative models outperformed RuTermExtract in terms of BERTScore (+4.78\% -- medical, +3.06\% -- linguistic, +2.5\% -- Math\&CS, +2.22\% -- historical). mBART showed the highest ROUGE-1 scores (8.96\% -- medical, 3.78\% - linguistic, 3.08\% -- Math\&CS, +2.68\% -- historical). Besides, ruGPT and mT5 outperformed RuTermExtract in terms of ROUGE-1 for the medical domain. In half of the cases, generative models showed a higher F1-score than RuTermExtract. But at the same time, mBART demonstrated superior F1-score for all domains (+12.16\% in comparison with RuTermExtract -- medical, +7.24\% -- linguistic, +5.83\% -- Math\&CS, +4.85\% -- historical). In general, mBART achieved the best scores across all considered domains. 

\begin{table}[h!]
\centering
\caption{In-domain results, \%. BERTScore -- BS, ROUGE-1 - R1, F1-score -- F1}
\begin{tabular}{|l|l|l|l|l|}\hline
Metric & Math\&CS & Historical & Medical & Linguistic \\\hline

\multicolumn{5}{|c|}{\textbf{Best baseline (RuTermExtract)}} \\\hline
BS & 75.95 & 73.53 & 74.93 & 74.15 \\\hline
R1 & 26.49 & 20.54 & 22.86 & 23.09 \\\hline
F1 & 11.02 & 9.65 & 11.35 & 11.19 \\\hline\hline

\multicolumn{5}{|c|}{\textbf{ruT5}} \\\hline
BS & 75.31 & 72.67 & \textbf{75.74} & 73.34 \\\hline
R1 & 20.87 & 12.95 & 20.43 & 13.5 \\\hline
F1 & 8.9 & 6 & \textbf{12.34} & 6.93 \\\hline\hline
\multicolumn{5}{|c|}{\textbf{ruGPT}} \\\hline
BS & \textbf{76.42} & \textbf{74.47} & \textbf{77.3} & \textbf{75.58} \\\hline
R1 & 22.82 & 18.7 & \textbf{24.63} & 21.73 \\\hline
F1 & 10.31 & \textbf{9.9} & \textbf{16.72} & \textbf{12.02} \\\hline\hline
\multicolumn{5}{|c|}{\textbf{mT5}} \\\hline
BS & \textbf{76.07} & \textbf{73.6} & \textbf{77.4} & \textbf{74.33} \\\hline
R1 & 24.79 & 16.04 & \textbf{24.19} & 19.54 \\\hline
F1 & \textbf{13.41} & 8.88 & \textbf{17.19} & 11.19 \\\hline\hline
\multicolumn{5}{|c|}{\textbf{mBART}} \\\hline
BS & \underline{\textbf{78.45}} & \underline{\textbf{75.75}} & \underline{\textbf{79.71}} & \underline{\textbf{77.21}} \\\hline
R1 & \underline{\textbf{29.57}} & \underline{\textbf{23.22}} & \underline{\textbf{31.82}} & \underline{\textbf{26.87}} \\\hline
F1 & \underline{\textbf{16.85}} & \underline{\textbf{14.5}} & \underline{\textbf{23.51}} & \underline{\textbf{18.43}}\\\hline
\end{tabular}
\label{table_in_domain_results}
\end{table}

Table \ref{table_evaluation_results} presents the main characteristics of generated keyphrases, namely, the average number of generated keyphrases per text and abstractness. Abstractness shows the proportion of generated keyphrases that do not appear in the corresponding source text.

\begin{table}[h!]
\centering
\caption{Evaluating generated keyphrases}
\begin{tabular}{|l|l|l|l|l|}\hline
Characteristic & Math\&CS & Historical & Medical & Linguistic \\\hline
\multicolumn{5}{|c|}{\textbf{ruT5}} \\\hline
Avg number of generated keyphrases & 5.25$\pm$2.12 & 4.91$\pm$2.27 & 3.9$\pm$1.52 & 2.15$\pm$0.37 \\\hline
Abstractness, \% & 61.23 & 74.05 & 63.03 & 68.89 \\\hline\hline
\multicolumn{5}{|c|}{\textbf{ruGPT}} \\\hline
Avg number of generated keyphrases & 4.33$\pm$1.54 & 4.6$\pm$1.66 & 4.2$\pm$1.34 & 4.85$\pm$1.57 \\\hline
Abstractness, \% & 75.68 & 67.88 & 57.39 & 61.88 \\\hline\hline
\multicolumn{5}{|c|}{\textbf{mT5}} \\\hline
Avg number of generated keyphrases & 3.65$\pm$1.36 & 4.44$\pm$2.12 & 3.52$\pm$1.1 & 6.05$\pm$2.64 \\\hline
Abstractness, \% & 48.72 & 71.27 & 56.33 & 69.46 \\\hline\hline
\multicolumn{5}{|c|}{\textbf{mBART}} \\\hline
Avg number of generated keyphrases & 4.07$\pm$1.17 & 4.75$\pm$1.57 & 4.03$\pm$1.26 & 5.02$\pm$1.32 \\\hline
Abstractness, \% & 38.72 & 49.6 & 36.13 & 42.56 \\\hline
\end{tabular}
\label{table_evaluation_results}
\end{table}

As can be seen from the table, ruGPT and mBART more accurately preserved the average number of keyphrases per text from the train set (see Table \ref{table_data}). mBART also showed less abstractness (from 36.13\% to 49.6\% for different domains) in comparison with other generative models. In some other cases, the models demonstrated high abstractness, for example, ruGPT -- 75.68\% for Math\&CS, ruT5 -- 74.05\% and mT5 -- 71.27\% for the historical domain. In most cases, ruT5, ruGPT, and mT5 showed higher abstractness than the proportion of absent keyphrases for the corresponding domain. On the contrary, mBART demonstrated less abstractness in comparison with the proportion of absent keyphrases.

Some examples of produced keyphrases are shown in Appendix \ref{appendix} (examples are given in Russian). Summarizing these examples, the main drawbacks of RuTermExtract are grammatical errors related to rule-based normalization and a large number of generic keywords that do not describe the specific topic of the text. Generative models showed good grammatical coherence and produced more specialized keyphrases. However, in one example, mT5 generated a non-existent word. The presented examples confirmed that mBART demonstrates less abstraction compared to other generative models.

The cross-domain results (\textbf{RQ2}) are provided in Table \ref{table_cross_domain_results}. The values that outperformed the corresponding scores achieved by baselines are shown in bold. As expected, the cross-domain scores were lower than the in-domain results. The most noticeable performance reduction in cross-domain evaluation was demonstrated by mT5. However, for some cases, the results obtained during cross-domain evaluation were still higher than baseline results. In particular, mBART demonstrated relatively high cross-domain performance in terms of BERTScore and F1-score for most domains.

\begin{table}[h!]
\centering
\caption{Cross-domain results, \%. BERTScore -- BS, ROUGE-1 - R1, F1-score -- F1; Math -- Math\&CS, Hist -- Historical, Med -- Medical, Ling -- Linguistic.}
\begin{tabular}{|l|l|l|l|l|l|l|l|l|l|l|l|l|}
\hline Train & \multicolumn{3}{c|}{Math\&CS} & \multicolumn{3}{c|}{Historical} & \multicolumn{3}{c|}{Medical} & \multicolumn{3}{c|}{Linguistic} \\\hline
Test & Hist & Med & Ling & Math & Med & Ling & Math & Hist & Ling & Math & Hist & Med \\\hline
\multicolumn{13}{|c|}{\textbf{ruT5}} \\\hline
BS & 71.62 & 74.79 & 73.98 & 73.29 & 71.64 & 72.62 & 73.91 & 71.59 & 73.07 & 73.49 & 71.29 & 72.43 \\
R1 & 11.58 & 18.17 & 17.03 & 14.47 & 9.61 & 12.67 & 17.03 & 11.66 & 13.84 & 14.26 & 9.34 & 9.41 \\
F1 & 4.3 & \textbf{11.61} & 6.28 & 4.81 & 4.27 & 5.76 & 6.28 & 4.46 & 6.79 & 4.25 & 3.15 & 3.51 \\
 \hline\hline
\multicolumn{13}{|c|}{\textbf{ruGPT}} \\\hline
BS & 72.45 & 74.99 & \textbf{74.13} & 75.35 & \textbf{74.95} & \textbf{76.07} & 74.87 & 72.61 & 73.95 & 75.17 & 73.22 & 74.68 \\
R1 & 12.74 & 17.01 & 16.04 & 19.67 & 18.12 & 17.6 & 18.53 & 13.22 & 15.81 & 19.96 & 14.59 & 18.26 \\
F1 & 5.04 & 9.21 & 7.04 & 8.09 & 10.46 & 8.43 & 7.25 & 5.91 & 7.82 & 7.37 & 6.07 & 9.23\\\hline\hline

\multicolumn{13}{|c|}{\textbf{mT5}} \\\hline
BS & 68.92 & 61.99 & 60.78 & 70.84 & 68.05 & 71.59 & 71.82 & 68.16 & 69.44 & 72.01 & 70.51 & 68.14 \\
R1 & 7.06 & 9.44 & 11.17 & 9.15 & 3.31 & 9.22 & 9.72 & 5 & 5.45 & 10.69 & 9.07 & 3.74 \\
F1 & 2.39 & 3.97 & 4.11 & 2.77 & 1.27 & 3.99 & 3.24 & 1.5 & 1.61 & 3.33 & 2.67 & 0.59 \\
 \hline\hline

\multicolumn{13}{|c|}{\textbf{mBART}} \\\hline
BS & 72.64 & \textbf{77.03} & \textbf{75.36} & \textbf{76.43} & \textbf{76.19} & \textbf{75.25} & \textbf{76.88} & \textbf{73.45} & \textbf{75.28} & \textbf{77.02} & \textbf{74.53} & \textbf{76.3} \\
R1 & 14.19 & \textbf{23.26} & 20.48 & 25.06 & 21.49 & 21.22 & 24.97 & 16.38 & 20.98 & 25.69 & 18.53 & 21.3 \\
F1 & 6.56 & \textbf{14.33} & \textbf{11.21} & \textbf{11.61} & \textbf{12.14} & \textbf{12.43} & \textbf{12.03} & 7.9 & \textbf{12.79} & \textbf{12.01} & 9.07 & \textbf{11.56} \\ \hline
\end{tabular}
\label{table_cross_domain_results}
\end{table}

\section{Conclusion}\label{conclusion}

In this work, we explored the effectiveness of fine-tuned generative transformer-based models in the task of keyphrase selection within Russian scientific texts. We described the results for generating lists of keyphrases and compared the performance of generative models with the performance of several unsupervised keyphrase extraction methods. In our experiments, generative models often demonstrated quality exceeding baselines. Moreover, the best results across all metrics and domains were achieved using the mBART model. The performance of generative models in cross-domain settings was expectedly lower. However, in several cases, cross-domain models also outperformed the baselines. The possible advantages of generating keyphrases using pre-trained language models is the absence of the need for setting restrictions on the number and length of keyphrases. Generative models can also produce keyphrases that are not explicitly mentioned in the original text.  

Future research could explore several promising directions. Firstly, further investigation into refining generative models' performance in cross-domain settings is warranted, as our study identified challenges in this area. Exploring novel approaches to leverage contextual information and linguistic features could also potentially enhance keyphrase selection accuracy. Furthermore, the possibility of generating the number of keyphrases specified by the user and setting other restrictions on the list of keyphrases generated by the model can also be investigated. From a methodological standpoint, exploring state-of-the-art instruction-based language models for keyphrase generation for Russian texts is a relevant research task. In this area, future research may focus on comparing classical keyphrase extraction approaches with those based on the use of fine-tuned and instruct-based language models.

%
%
%
\bibliographystyle{splncs04}
\bibliography{samplepaper.bib}

\begin{thebibliography}{10}
\providecommand{\url}[1]{\texttt{#1}}
\providecommand{\urlprefix}{URL }
\providecommand{\doi}[1]{https://doi.org/#1}

\bibitem{bird2006nltk}
Bird, S.: {NLTK}: the natural language toolkit. In: Proceedings of the COLING/ACL 2006 Interactive Presentation Sessions. pp. 69--72 (2006)

\bibitem{bougouin2013topicrank}
Bougouin, A., Boudin, F., Daille, B.: Topic{R}ank: Graph-based topic ranking for keyphrase extraction. In: International joint conference on natural language processing (IJCNLP). pp. 543--551 (2013)

\bibitem{campos2020yake}
Campos, R., Mangaravite, V., Pasquali, A., Jorge, A., Nunes, C., Jatowt, A.: {YAKE}! keyword extraction from single documents using multiple local features. Information Sciences  \textbf{509},  257--289 (2020). \doi{10.1016/j.ins.2019.09.013}

\bibitem{ccano2019keyphrase}
{\c{C}}ano, E., Bojar, O.: Keyphrase generation: A multi-aspect survey. In: 2019 25th Conference of Open Innovations Association (FRUCT). pp. 85--94. IEEE (2019). \doi{10.23919/FRUCT48121.2019.8981519}

\bibitem{cano2019keyphrase}
Cano, E., Bojar, O.: Keyphrase generation: A text summarization struggle. In: Proceedings of the 2019 Conference of the North American Chapter of the Association for Computational Linguistics: Human Language Technologies, Volume 1 (Long and Short Papers). pp. 666--672 (2019). \doi{10.18653/v1/N19-1070}

\bibitem{celikyilmaz2020evaluation}
Celikyilmaz, A., Clark, E., Gao, J.: Evaluation of text generation: A survey. arXiv preprint arXiv:2006.14799  (2020)

\bibitem{chan2019neural}
Chan, H.P., Chen, W., Wang, L., King, I.: Neural keyphrase generation via reinforcement learning with adaptive rewards. In: Proceedings of the 57th Annual Meeting of the Association for Computational Linguistics. pp. 2163--2174 (2019). \doi{10.18653/v1/P19-1208}

\bibitem{chen2023enhancing}
Chen, B., Iwaihara, M.: Enhancing keyphrase generation by {BART}finetuning with splitting and shuffling. In: Pacific Rim International Conference on Artificial Intelligence. pp. 305--310. Springer (2023). \doi{10.1007/978-981-99-7019-3_29}

\bibitem{chen2018keyphrase}
Chen, J., Zhang, X., Wu, Y., Yan, Z., Li, Z.: Keyphrase generation with correlation constraints. In: Proceedings of the 2018 Conference on Empirical Methods in Natural Language Processing. pp. 4057--4066 (2018). \doi{10.18653/v1/D18-1439}

\bibitem{devlin-etal-2019-bert}
Devlin, J., Chang, M.W., Lee, K., Toutanova, K.: {BERT}: Pre-training of deep bidirectional transformers for language understanding. In: 2019 Conference of the North {A}merican Chapter of the Association for Computational Linguistics: Human Language Technologies, Volume 1 (Long and Short Papers). pp. 4171--4186. Association for Computational Linguistics, Minneapolis, Minnesota (2019). \doi{10.18653/v1/N19-1423}

\bibitem{glazkova2023keyphrase}
Glazkova, A.V., Morozov, D.A., Vorobeva, M.S., Stupnikov, A.A.: Keyphrase generation for the {R}ussian-language scientific texts using m{T}5. Modelirovanie i Analiz Informatsionnykh Sistem  \textbf{30}(4),  418--428 (2023). \doi{10.18255/1818-1015-2023-4-418-428}

\bibitem{grootendorst2020keybert}
Grootendorst, M.: Key{BERT}: Minimal keyword extraction with bert. (2020). \doi{10.5281/zenodo.4461265}, \url{https://doi.org/10.5281/zenodo.4461265}

\bibitem{guseva2024}
Guseva, D., Mitrofanova, O.: Keyphrases in {R}ussian-language popular science texts: comparison of oral and written speech perception with the results of automatic analysis. Terra Linguistica  \textbf{15}(1),  20--35 (2024). \doi{10.18721/JHSS.15102}

\bibitem{jiang2023generating}
Jiang, Y., Meng, R., Huang, Y., Lu, W., Liu, J.: Generating keyphrases for readers: A controllable keyphrase generation framework. Journal of the Association for Information Science and Technology  \textbf{74}(7),  759--774 (2023). \doi{10.1002/asi.24749}

\bibitem{koloski2021extending}
Koloski, B., Pollak, S., {\v{S}}krlj, B., Martinc, M.: Extending neural keyword extraction with {TF-IDF} tagset matching. In: Proceedings of the EACL Hackashop on News Media Content Analysis and Automated Report Generation. pp. 22--29 (2021)

\bibitem{korobov2015morphological}
Korobov, M.: Morphological analyzer and generator for {R}ussian and {U}krainian languages. In: Analysis of Images, Social Networks and Texts: 4th International Conference, AIST 2015, Yekaterinburg, Russia, April 9--11, 2015, Revised Selected Papers 4. pp. 320--332. Springer (2015). \doi{10.1007/978-3-319-26123-2_31}

\bibitem{kulkarni2022learning}
Kulkarni, M., Mahata, D., Arora, R., Bhowmik, R.: Learning rich representation of keyphrases from text. In: Findings of the Association for Computational Linguistics: NAACL 2022. pp. 891--906 (2022). \doi{10.18653/v1/2022.findings-naacl.67}

\bibitem{kuratov2019adaptation}
Kuratov, Y., Arkhipov, M.: Adaptation of deep bidirectional multilingual transformers for {R}ussian language. In: Komp'juternaja Lingvistika i Intellektual'nye Tehnologii. pp. 333--339 (2019)

\bibitem{lewis2020bart}
Lewis, M., Liu, Y., Goyal, N., Ghazvininejad, M., Mohamed, A., Levy, O., Stoyanov, V., Zettlemoyer, L.: {BART}: Denoising sequence-to-sequence pre-training for natural language generation, translation, and comprehension. In: Proceedings of the 58th Annual Meeting of the Association for Computational Linguistics. pp. 7871--7880 (2020). \doi{10.18653/v1/2020.acl-main.703}

\bibitem{lin2004rouge}
Lin, C.Y.: {ROUGE}: A package for automatic evaluation of summaries. In: Text summarization branches out. pp. 74--81 (2004)

\bibitem{liu2019roberta}
Liu, Y., Ott, M., Goyal, N., Du, J., Joshi, M., Chen, D., Levy, O., Lewis, M., Zettlemoyer, L., Stoyanov, V.: Roberta: A robustly optimized bert pretraining approach. arXiv preprint arXiv:1907.11692  (2019)

\bibitem{martinez2023chatgpt}
Mart{\'\i}nez-Cruz, R., L{\'o}pez-L{\'o}pez, A.J., Portela, J.: Chat{GPT} vs state-of-the-art models: a benchmarking study in keyphrase generation task. arXiv preprint arXiv:2304.14177  (2023)

\bibitem{meng2017deep}
Meng, R., Zhao, S., Han, S., He, D., Brusilovsky, P., Chi, Y.: Deep keyphrase generation. In: Proceedings of the 55th Annual Meeting of the Association for Computational Linguistics (Volume 1: Long Papers). pp. 582--592 (2017). \doi{10.18653/v1/P17-1054}

\bibitem{mitrofanova2022experiments}
Mitrofanova, O., Gavrilic, D.: Experiments on automatic keyphrase extraction in stylistically heterogeneous corpus of {R}ussian texts. Terra Linguistica  \textbf{50}(4),  22--40 (2022). \doi{10.18721/JHSS.13402}

\bibitem{morozov2022dataset}
Morozov, D., Glazkova, A.: Keyphrases {CS\&M}ath {R}ussian (2022). \doi{10.17632/dv3j9wc59v.1}, \url{https://data.mendeley.com/datasets/dv3j9wc59v/1}

\bibitem{morozov2023generation}
Morozov, D., Glazkova, A., Tyutyulnikov, M., Iomdin, B.: Keyphrase generation for abstracts of the {R}ussian-language scientific articles. NSU Vestnik. Series: Linguistics and Intercultural Communication  \textbf{21}(1),  54--66 (2023). \doi{10.25205/1818-7935-2023-21-1-54-66}

\bibitem{mu2020keyphrase}
Mu, F., Yu, Z., Wang, L., Wang, Y., Yin, Q., Sun, Y., Liu, L., Ma, T., Tang, J., Zhou, X.: Keyphrase extraction with span-based feature representations. arXiv preprint arXiv:2002.05407  (2020)

\bibitem{narin2021content}
Narin, N.G.: A content analysis of the metaverse articles. Journal of Metaverse  \textbf{1}(1),  17--24 (2021)

\bibitem{nguyen2021keyphrase}
Nguyen, Q.H., Zaslavskiy, M.: Keyphrase extraction in {R}ussian and {E}nglish scientific articles using sentence embeddings. In: 2021 28th Conference of Open Innovations Association (FRUCT). pp.~1--7. IEEE (2021). \doi{10.23919/FRUCT50888.2021.9347584}

\bibitem{papagiannopoulou2020review}
Papagiannopoulou, E., Tsoumakas, G.: A review of keyphrase extraction. Wiley Interdisciplinary Reviews: Data Mining and Knowledge Discovery  \textbf{10}(2),  e1339 (2020). \doi{10.1002/widm.1339}

\bibitem{raffel2020exploring}
Raffel, C., Shazeer, N., Roberts, A., Lee, K., Narang, S., Matena, M., Zhou, Y., Li, W., Liu, P.J.: Exploring the limits of transfer learning with a unified text-to-text transformer. Journal of machine learning research  \textbf{21}(140),  1--67 (2020)

\bibitem{rose2010automatic}
Rose, S., Engel, D., Cramer, N., Cowley, W.: Automatic keyword extraction from individual documents. Text mining: applications and theory pp. 1--20 (2010). \doi{10.1002/9780470689646.ch1}

\bibitem{sahrawat2020keyphrase}
Sahrawat, D., Mahata, D., Zhang, H., Kulkarni, M., Sharma, A., Gosangi, R., Stent, A., Kumar, Y., Shah, R.R., Zimmermann, R.: Keyphrase extraction as sequence labeling using contextualized embeddings. In: Advances in Information Retrieval: 42nd European Conference on IR Research, ECIR 2020, Lisbon, Portugal, April 14--17, 2020, Proceedings, Part II 42. pp. 328--335. Springer (2020). \doi{10.1007/978-3-030-45442-5_41}

\bibitem{sandul2018keyword}
Sandul, M.V., Mikhailova, E.G.: Keyword extraction from single {R}ussian document. In: Proceedings of the Third Conference on Software Engineering and Information Management. pp. 30--36 (2018)

\bibitem{rutermextract}
Shevchenko, I.: {R}u{T}erm{E}xtract. \url{https://github.com/igor-shevchenko/rutermextract} (2018)

\bibitem{sokolova2017automatic}
Sokolova, E., Mitrofanova, O.: Automatic keyphrase extraction by applying {KEA} to {R}ussian texts. In: IMS (CLCO). pp. 157--165 (2017). \doi{10.17586/2541-9781-2017-1-157-165}

\bibitem{sokolova2018keyphrase}
Sokolova, E., Moskvina, A., Mitrofanova, O.: Keyphrase extraction from the {R}ussian corpus on linguistics by means of {KEA} and {RAKE} algorithms. In: Data Analytics and Management in Data Intensive Domains. pp. 369--372 (2018)

\bibitem{song2022hyperbolic}
Song, M., Feng, Y., Jing, L.: Hyperbolic relevance matching for neural keyphrase extraction. In: Proceedings of the 2022 Conference of the North American Chapter of the Association for Computational Linguistics: Human Language Technologies. pp. 5710--5720 (2022). \doi{10.18653/v1/2022.naacl-main.419}

\bibitem{song2023survey}
Song, M., Feng, Y., Jing, L.: A survey on recent advances in keyphrase extraction from pre-trained language models. Findings of the Association for Computational Linguistics: EACL 2023 pp. 2153--2164 (2023). \doi{10.18653/v1/2023.findings-eacl.161}

\bibitem{song2023chatgpt}
Song, M., Jiang, H., Shi, S., Yao, S., Lu, S., Feng, Y., Liu, H., Jing, L.: Is {ChatGPT} a good keyphrase generator? a preliminary study. arXiv preprint arXiv:2303.13001  (2023)

\bibitem{sun2020joint}
Sun, S., Xiong, C., Liu, Z., Liu, Z., Bao, J.: Joint keyphrase chunking and salience ranking with {BERT}. arXiv preprint arXiv:2004.13639  (2020)

\bibitem{swaminathan2020preliminary}
Swaminathan, A., Zhang, H., Mahata, D., Gosangi, R., Shah, R., Stent, A.: A preliminary exploration of gans for keyphrase generation. In: Proceedings of the 2020 Conference on Empirical Methods in Natural Language Processing (EMNLP). pp. 8021--8030 (2020). \doi{10.18653/v1/2020.emnlp-main.645}

\bibitem{tang2020multilingual}
Tang, Y., Tran, C., Li, X., Chen, P.J., Goyal, N., Chaudhary, V., Gu, J., Fan, A.: Multilingual translation with extensible multilingual pretraining and finetuning. arXiv preprint arXiv:2008.00401  (2020)

\bibitem{widyassari2022review}
Widyassari, A.P., Rustad, S., Shidik, G.F., Noersasongko, E., Syukur, A., Affandy, A., et~al.: Review of automatic text summarization techniques \& methods. Journal of King Saud University-Computer and Information Sciences  \textbf{34}(4),  1029--1046 (2022). \doi{10.1016/j.jksuci.2020.05.006}

\bibitem{wienecke2020automatic}
Wienecke, Y.: Automatic keyphrase extraction from {R}ussian-language scholarly papers in computational linguistics. University Honors Theses  (2020). \doi{10.15760/honors.957}

\bibitem{witten1999kea}
Witten, I.H., Paynter, G.W., Frank, E., Gutwin, C., Nevill-Manning, C.G.: Kea: Practical automatic keyphrase extraction. In: Proceedings of the fourth ACM conference on Digital libraries. pp. 254--255 (1999)

\bibitem{wu2022pre}
Wu, D., Ahmad, W.U., Chang, K.W.: Pre-trained language models for keyphrase generation: A thorough empirical study. arXiv preprint arXiv:2212.10233  (2022)

\bibitem{xue2021mt5}
Xue, L., Constant, N., Roberts, A., Kale, M., Al-Rfou, R., Siddhant, A., Barua, A., Raffel, C.: m{T5}: A massively multilingual pre-trained text-to-text transformer. In: Proceedings of the 2021 Conference of the North American Chapter of the Association for Computational Linguistics: Human Language Technologies. pp. 483--498 (2021). \doi{10.18653/v1/2021.naacl-main.41}

\bibitem{zhangbertscore}
Zhang, T., Kishore, V., Wu, F., Weinberger, K.Q., Artzi, Y.: {BERTS}core: Evaluating text generation with {BERT}. In: International Conference on Learning Representations

\bibitem{zhang2018keyphrase}
Zhang, Y., Xiao, W.: Keyphrase generation based on deep seq2seq model. IEEE access  \textbf{6},  46047--46057 (2018). \doi{10.1109/ACCESS.2018.2865589}

\bibitem{zmitrovich2023family}
Zmitrovich, D., Abramov, A., Kalmykov, A., Tikhonova, M., Taktasheva, E., Astafurov, D., Baushenko, M., Snegirev, A., Shavrina, T., Markov, S., et~al.: A family of pretrained transformer language models for {R}ussian. arXiv preprint arXiv:2309.10931  (2023)

\end{thebibliography}

\newpage
\appendix
\section{Appendix}\label{appendix}

Table \ref{table7} contains two examples of keyphrases generated using different models. The keyphrases that match the keyphrases from the original list (without normalization) are highlighted in bold. The tokens (words) that match the tokens from the original list are underlined. The tokens that that do not appear in the abstract are double underlined. The keyphrases with grammatical or spelling errors are italicized.

\begin{table}[h!]
\tiny
\caption{The examples of produced keyphrases. Original -- original (author's) list of keyphrases, RTE -- RuTermExtract.}
\begin{tabular}{|p{1cm}|p{11cm}|}\hline
\label{table7}
\scriptsize
Method &Keyphrases \\\hline
\multicolumn{2}{|p{12cm}|}{\textbf{Abstract 1.} \selectlanguage{russian}Данная статья посвящена проекту создания межкафедральной виртуальной лаборатории на базе лабораторного стенда с универсальной платой АЦП/ЦАП, а также смоделированных приборов в средах Matlab, LabView \textit{(Istomin, V. V. (2010). Organization of virtual education in branches 200300 and 200400. Izvestiya SFedU. Engineering Sciences).\selectlanguage{british}}} \\\hline
\selectlanguage{british}Original & \selectlanguage{russian}виртуальная лаборатория, удаленное обучение, моделирование приборов, лабораторный стенд, лабораторный практикум\\\hline
RTE & \selectlanguage{russian}межкафедральная \underline{виртуальная лаборатория}, универсальная плата, среды matlab, смоделированные \underline{приборы}, проект создания, \textit{\textbf{лабораторное стенд}}, данная статья, база, ацп/цап, labview\selectlanguage{british}\\\hline
ruT5 & \selectlanguage{russian}\textbf{виртуальные лаборатории}, \textbf{лабораторный стенд}\selectlanguage{british} \\\hline
ruGPT & \selectlanguage{russian}\textbf{лабораторный стенд}, \underline{моделирование}, \underline{\underline{программный комплекс}}, \underline{\underline{вычислительный эксперимент}}, \underline{\underline{интеграция, программная платформа}}, \underline{\underline{численная реализация}}\selectlanguage{british}\\\hline
mT5 & \selectlanguage{russian}\textbf{лабораторный стенд}, \textbf{виртуальная лаборатория}, \underline{\underline{исследовательский}} \underline{стенд}, \textit{\underline{\underline{межкавказальная}}} \textbf{виртуальная лаборатория}\selectlanguage{british}\\\hline
mBART & \selectlanguage{russian}межкафедральная \underline{виртуальная лаборатория}, \underline{виртуальный} \underline{прибор}, \textbf{лабораторный стенд}, универсальная плата \selectlanguage{british}\\\hline

\multicolumn{2}{|p{12cm}|}{\textbf{Abstract 2.} \selectlanguage{russian}В работе рассмотрены проблемы интеграции разнородных геоинформационных ресурсов и создания единой базы данных, необходимой для разработки геоинформационного портала отрасли. Описан подход к отображению семантики, заложенной в пространственные онтологии, в географические концептуальные схемы для представления хранимой в базах геоданных информации \textit{(Duhin, S. V. (2006). Designing a geoinformation portal for the industry using geodata ontologies. Upravlenie bol'shimi sistemami: sbornik trudov.}}  \\\hline
Original & \selectlanguage{russian}геоинформационные системы, геоинформационный портал, онтологии, географические концептуальные модели\\\hline
RTE & \selectlanguage{russian}разнородные \underline{геоинформационные} ресурсы, единая база данных, \underline{геоинформационный портал} отрасли, \textit{\underline{географические концептуальные} схема}, базы геоданных информации, \textit{пространственные \underline{онтология}}, проблемы интеграции, отображение семантики, создание, разработка\selectlanguage{british}\\\hline
ruT5 & \selectlanguage{russian}\underline{геоинформационные} ресурсы, \underline{геоинформационная} база данных, \underline{геоинформационные} \underline{\underline{средства}}, \textbf{геоинформационный портал}, \underline{геоинформационный} ресурс отрасли, \textbf{геоинформационные \underline{\underline{системы}}}\selectlanguage{british} \\\hline
ruGPT & \selectlanguage{russian}база данных, \textbf{онтологии}, \textbf{геоинформационный портал}\selectlanguage{british}\\\hline
mT5 & \selectlanguage{russian}\textbf{геоинформационные \underline{\underline{системы}}}, пространственные \underline{онтологии}, \underline{концептуальные} схемы, семантическая интеграция, интеграция данных, единое \underline{\underline{информационное пространство}} \selectlanguage{british}\\\hline
mBART & \selectlanguage{russian}\underline{\underline{информационные технологии}}, \textbf{геоинформационные \underline{\underline{системы}}}, пространственная \underline{онтология}\selectlanguage{british}\\\hline
\end{tabular}
\end{table}

\end{document}